\documentclass{article}

\usepackage{PRIMEarxiv}

\usepackage[utf8]{inputenc} 
\usepackage[T1]{fontenc}    
\usepackage{hyperref}       
\usepackage{url}            
\usepackage{booktabs}       
\usepackage{amsfonts}       
\usepackage{nicefrac}       
\usepackage{microtype}      
\usepackage{lipsum}
\usepackage{fancyhdr}       
\usepackage{graphicx}       

\usepackage{multirow}

\graphicspath{{media/}}     

\title{Context based lemmatizer for Polish language
\thanks{\textit{\underline{Citation}}: 
\textbf{Authors. Title. Pages.... DOI:000000/11111.}} 
}

\author{
  Michal Karwatowski \\
  UST-AGH \\
  Cracow\\
  \texttt{mkarwat@agh.edu.pl} \\
   \And
  Marcin Pietron \\
  UST-AGH \\
  Cracow\\
  \texttt{pietron@agh.edu.pl} \\
}

\begin{document}
\maketitle

\begin{abstract}
Lemmatization is the process of grouping together the inflected forms of a word so they can be analysed as a single item, identified by the word's lemma, or dictionary form. In computational linguistics, lemmatisation is the algorithmic process of determining the lemma of a word based on its intended meaning. Unlike stemming, lemmatisation depends on correctly identifying the intended part of speech and meaning of a word in a sentence, as well as within the larger context surrounding that sentence. As a result, developing efficient lemmatisation algorithm is the complex task. In recent years it can be observed that deep learning models used for this task outperform other methods including machine learning algorithms. In this paper the polish lemmatizer based on Google T5 model is presented. The training was run with different context lengths. The model achieves the best results for polish language lemmatisation process.
\end{abstract}


\section{Introduction}

Natural Language Processing consists of many tasks, the role of each is extracting and processing human understandable meaning from the text data. Some tasks like classification encompass the complete flow from data to answer, in other tasks like part of speech tagging, results are often used as an input for next algorithms. An interesting and complex problem is translation, where the meaning of the text needs to be extracted and encoded back to the text in a different language. This approach describes a family of NLP tasks called text-to-text or sequence-to-sequence processing.
Another example of text-to-text processing is lemmatisation, it finds a base form of a given word or expression. Complexity of this problem varies from language to language. In English the number of word variations is usually low, there are simple rules and not many exceptions. However in Slavic languages such as Polish inflection of words it is significantly more complicated and effective lemmatisation is beyond capabilities of a rule based or edit tree classification methods \cite{chakrabarty2017}, \cite{straka2016}. Situation becomes more difficult when we include multi-segment expressions. For example: “ministerstwie kultury i dziedzictwa narodowego” should be lemmatized to “ministerstwo kultury i dziedzictwa narodowego” not “ministerstwo kultura i dziedzictwo narodowy”. In such cases context and inflection of words should be taken for correct lemmatisation. Another issue are foreign phrases changed using Polish grammar, which is especially common with names and places. The next problem in lemmatisation arise from encountering previously unseen words during inference time. In recent years it can be observed that sequence-to-sequence models used for lemmatisation of highly inflective languages outperform other methods \cite{kanerva2019}. In this paper the Google T5 model was trained on polish language dataset for PolEval competition. The training was run with three configurations: no context, fixed and variable context lengths. The experiments show the efficiency of fine-tuned Google T5 based lemmatiser model on polish language and the impact the context on model accuracy. The context based Google T5 model achieves best results for polish language which were measured so far. The simulations were run with small and base T5 models.   

\section{Related work}
\label{sec:related}

There were a few approaches to Polish language lemmatisation, for example KRNNT tagger \cite{wrobel2017} based on recurrent neural network. In 2019 lemmatisation was a part of a PolEval competition\footnote{http://2019.poleval.pl/index.php/tasks/task2}. The most common machine learning approaches for lemmatisation are based on edit tree classification, where all possible edit trees or word-to-lemma transformation rules are gathered from the training data. Then a classifier is trained to choose the correct one for a given input word. Many recent works build on the sequence-to-sequence learning paradigm \cite{bergmanis2017}, \cite{bergmanis2018}, \cite{chakrabarty2017}. 
\cite{muller2017}, \cite{straka2018} and \cite{kondrat2015}. Bergmanis and Goldwater \cite{bergmanis2017} present the Lematus context-sensitive lemmatisation system, where the model is trained to generate the lemma from a given input word one character at a time. The Lematus system outperforms other context-aware lemmatisation systems. \cite{kanerva2019}
Sequence-to-sequence models have also been applied in the context of morphological reinflection, the reverse of the lemmatisation task. In the CoNLL-SIGMORPHON 2017 Shared Task on Universal Morphological Reinflection \cite{cott2017} several of the top-ranking systems were based on sequence-to-sequence learning \cite{kann2017}, \cite{bergmanis2017}.

\section{Model architecture and training setup}

\subsection{Text-to-text model}

Google T5 \cite{2020t5} is an extremely large new neural network model that is trained on a mixture of unlabeled text and labeled data from popular natural language processing tasks. Google's T5 is one of the most advanced natural language models to date. It builds on top of previous work on Transformer based models. Unlike BERT, which had only encoder blocks, and GPT-2, which had only decoder blocks, T5 uses both. This work utilises Google’s T5 text-to-text transformer, which is initially trained for English language, however Allegro research team\footnote{https://ml.allegro.tech/} retrained this model for Polish language and shared final models. Training this model for a new language is very time and resource consuming, but thanks to transfer learning technique it is enough to perform this only once, and afterwards fine tune it for a specific downstream task. The small and base versions of T5 model were used in experiments.

\subsection{Dataset}

First step required to train a model is to gather enough data of sufficiently high quality. We used datasets from both task 2 and 3 of PolEval’19 competition \footnote{http://2019.poleval.pl/}. Task 2 data is specific to lemmatisation however it is not very big. Therefore we added tagged data from entity linking task 3. The final dataset consists of 170k pairs of orthographic and lemmatized forms. Where 20k samples were from task 2 data, 100k were phrases from task 3, and 50k were single words from tasks 3.

\subsection{Additional context information}

First experiments were performed only on the specified words or phrases. Then the addition of context was explored. To provide context during training each sample in data set has context attached. Maximal context span was set to 64 words before and after selected phrase. If the phrase was near the beginning of document (or end) and the context could not be fully filled the remaining space was left empty. Additionally, phrases position within context was clearly marked to provide mechanism to extract only a portion of the context.

\subsection{Evaluation}

To evaluate performance of the model three accuracy metrics are utilised, each counting prediction as correct only if all characters are exactly matched to the ground truth:

\begin{itemize}
    \item AccCS - accuracy case sensitive, all characters in prediction are exactly the same
    \item AccCI - accuracy case insensitive, all characters in prediction are the same however, lowercase and uppercase letters are treated as equal
    \item Score - combination of the two, given by equation \ref{eq:score}.
\end{itemize}

\begin{equation}
\label{eq:score}
Score = 0.2 *AccCS+0.8*AccCI    
\end{equation}

\section{Experiments}

\subsection{Training with no context}

T5 model was retrained with different configurations to perform lemmatisation task. In first experiment context was not used, only the phrases that were to be lemmtized. That solution is consistent with the previous approaches. Final results are summarized in table \ref{tab:no-context}. Baseline configuration did not perform any computing, only passed input as an output, accuracy score around 50\% indicates that half of the input is already in its base form, and the model needs to recognise that no changes are required. Our text-to-text solution outperforms previous results significantly.

\begin{table}[]
\centering
\begin{tabular}{|c|c|c|c|}
\hline
\textbf{Model} & \textbf{AccCS} & \textbf{AccCI} & \textbf{Score} \\ \hline
PolEval'19 winner & 84,78 & 88,13 & 87,46 \\ \hline
Baseline          & 50,35 & 51,73 & 51,46 \\ \hline
KRNNT             & 65,26 & 70,76 & 69,66 \\ \hline
small (ours)   & 87,89 & 89,72 & 89,36 \\ \hline
base  (ours)   & 88,62 & 90,44 & 90,07 \\ \hline
large (ours)   & \textbf{89,33} & \textbf{91,23} & \textbf{90,85} \\ \hline
\end{tabular}
\caption{No context}
\label{tab:no-context}
\end{table}

\subsection{Fixed context}

Next experiment utilised additional context information. The hypothesis was that given more information about context in witch the phrase was used should allow the network to make better predictions \cite{bergmanis2017}, \cite{kanerva2019}. Both the phrase and the context are given at the input of the network, where first comes the phrase, then separation token and context, filled with padding tokens to match required input size. Table \ref{tab:fixed-context} summarizes the results, Additional context information improved the score by up to 1.9\% for base model size and 1\% for small version.

\begin{table}[]
\centering
\begin{tabular}{|c|c|c|c|c|}
\hline
\textbf{Model}         & \textbf{context} & \textbf{AccCS} & \textbf{AccCI} & \textbf{Score} \\ \hline
\multirow{2}{*}{small} & 16               & 88,96          & 90,67          & 90,33          \\ \cline{2-5} 
                       & 64               & 88,98          & 90,64          & 90,31          \\ \hline
\multirow{2}{*}{base}  & 16               & 90,58          & 92,31          & 91,97          \\ \cline{2-5} 
                       & 64               & 90,31          & 92,06          & 91,71          \\ \hline
\end{tabular}
\caption{Fixed context}
\label{tab:fixed-context}
\end{table}

\subsection{Variable context}

\begin{figure}
    \centering
    \includegraphics[width=0.75\hsize]{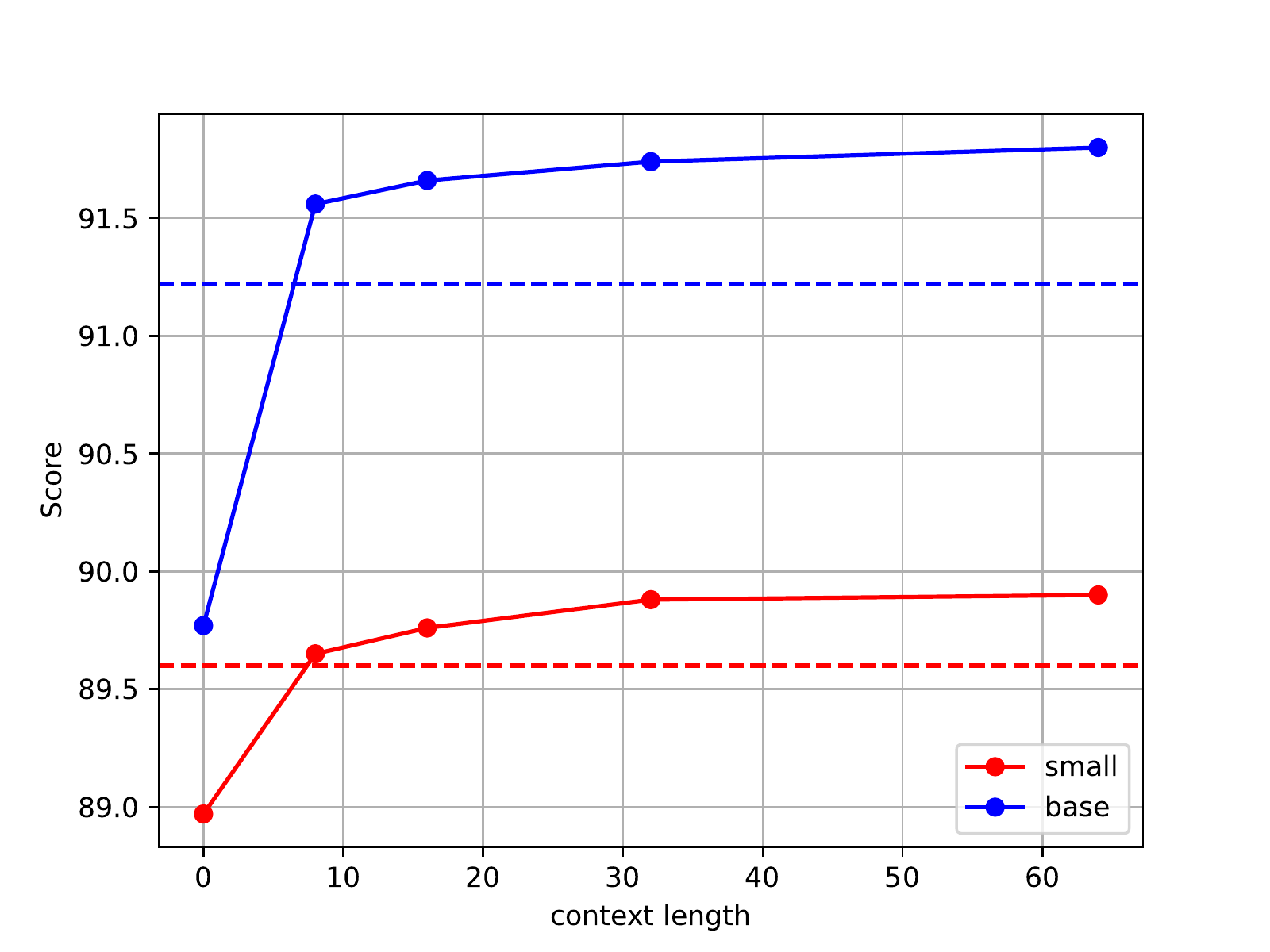}
    \caption{Lemmatizer score depending on context length, for network trained using variable context method. Dot points indicate tests with fixed context, dashed line indicate test with variable context.}
    \label{fig:lemma_context}
\end{figure}

While using fixed context size for prediction yields good results, it is not always practical to use. In many real world scenarios context size is not known in advance. For such situations we have trained a model with variable context length configuration. In training each sample had 30\% probability to be provided without context, for remaining samples context length was randomly chosen from 8 to 64 range with uniform distribution. To evaluate the model a series of tests were performed: a complete test set without context, range of contexts of fixed scope, and finally a test with the same variable context settings as during training. Final results are presented in Table \ref{fig:lemma_context}.

\begin{table}[]
\centering
\begin{tabular}{|c|c|r|r|r|}
\hline
\textbf{Model} & \textbf{context} & \multicolumn{1}{c|}{\textbf{AccCS}} & \multicolumn{1}{c|}{\textbf{AccCI}} & \multicolumn{1}{c|}{\textbf{Score}} \\ \hline
\multirow{6}{*}{small} & no       & 87,51          & 89,33          & 88,97          \\ \cline{2-5} 
                       & fixed 8  & 88,24          & 90,00          & 89,65          \\ \cline{2-5} 
                       & fixed 16 & 88,36          & 90,11          & 89,76          \\ \cline{2-5} 
                       & fixed 32 & 88,51          & 90,22          & 89,88          \\ \cline{2-5} 
                       & fixed 64 & \textbf{88,55} & \textbf{90,24} & \textbf{89,90} \\ \cline{2-5} 
                       & variable & 88,19          & 89,95          & 89,60          \\ \hline
\multirow{6}{*}{base}  & no       & 88,05          & 90,21          & 89,77          \\ \cline{2-5} 
                       & fixed 8  & 90,09          & 91,92          & 91,56          \\ \cline{2-5} 
                       & fixed 16 & 90,25          & 92,01          & 91,66          \\ \cline{2-5} 
                       & fixed 32 & 90,32          & 92,10          & 91,74          \\ \cline{2-5} 
                       & fixed 64 & \textbf{90,39} & \textbf{92,16} & \textbf{91,80} \\ \cline{2-5} 
                       & variable & 89,79          & 91,57          & 91,22          \\ \hline
\end{tabular}
\caption{Variable context}
\label{tab:context-variable}
\end{table}

This series of experiments allowed as to check how context length affects prediction accuracy (fig. \ref{fig:lemma_context}). It is visible that even small context increases accuracy significantly, and the greater the scope of the context the better results however, accuracy increase with context length is not as prominent as from no context to short context.

\section{Conclusion}
Presented experiments confirm the validity of text to text approach. Additional context information improved the prediction accuracy, also increase of context length correlated with the increase of accuracy however, even 8 words context was enough to rise the performance notably. Presented approach allowed to use one model regardless of context length or presence, facilitating practical use. The best model achieved $92.16\%$ case insensitive accuracy significantly outperforming previous solutions.

All models were trained on PLGrid infrastructure using Prometheus servers with NVIDIA V100 GPUs. Time required to train one epoch varied depending on model size, from 0.5h for small, 2.5h for base, to over 9h for large. Each configuration used 4 GPUs.

\section*{Acknowledgments}
This research was supported in part by PLGrid Infrastructure.

\bibliographystyle{unsrt}  
\bibliography{references}

\end{document}